\documentclass[11pt]{article}
\usepackage{eacl2017}
\usepackage{times}
\usepackage{url}
\usepackage{latexsym}
\usepackage{graphicx}
\usepackage[utf8]{inputenc}

\eaclfinalcopy 

\title{Cross-Lingual Dependency Parsing for Closely Related Languages -- Helsinki's Submission to VarDial 2017}

\author{Jörg Tiedemann\\
  Department of Modern Languages\\
  University of Helsinki\\
  {first.lastname@helsinki.fi}}

\date{}

\begin{document}
\maketitle
\begin{abstract}
This paper describes the submission from the University of Helsinki to the shared task on cross-lingual dependency parsing at VarDial 2017. We present work on annotation projection and treebank translation that gave good results for all three target languages in the test set. In particular, Slovak seems to work well with information coming from the Czech treebank, which is in line with related work. The attachment scores for cross-lingual models even surpass the fully supervised models trained on the target language treebank. Croatian is the most difficult language in the test set and the improvements over the baseline are rather modest. Norwegian works best with information coming from Swedish whereas Danish contributes surprisingly little.
\end{abstract}

\section{Introduction}
\label{intro}

Cross-lingual parsing is interesting as a cheap method for bootstrapping tools in a new language from resources in another language. Various approaches have been proposed in the literature, which can mainly be divided into data transfer (i.e. annotation projection, e.g. \cite{Hwa05bootstrappingparsers}) and model transfer approaches (e.g. delexicalized models such as \cite{mcdonald2013universal}). We will focus on data transfer in this paper using annotation projection and machine translation to transform source language treebanks to be used as training data for dependency parsers in the target language. Our previous work has shown that these techniques are quite robust and show better performance than simple transfer models based on delexicalized parsers \cite{TiedemannAgic:2016:JAIR}. This is especially true for real-world test cases in which part-of-speech (PoS) labels are predicted instead of given as gold standard annotation while testing the parsing models \cite{tiedemann:2015:Depling}.

Cross-lingual parsing assumes strong syntactic similarities between source and target language which can be seen at the degradation of model performance when using distant languages such as English and Finnish \cite{tiedemann:2015:NODALIDA}. The task at VarDial, therefore, focuses on closely related languages, which makes more sense also from a practical point of view. Many pools of closely related languages and language variants exist and, typically, the support in terms of resources and tools is very biased towards one of the languages in such a pool. Hence, one can say that the task at VarDial simulates real-world cases using existing resources from the universal dependencies project \cite{nivre2016universal} and promotes the ideas for practical application development. The results show that this test is, in fact, not only a simulation but actually improves the results for one of the languages in the test set: Slovak. Cross-lingual models outperform the supervised upper bound, which is a great result in favor of the transfer learning ideas.

More details about the shared task on cross-lingual parsing at VarDial 2017 can be found in \cite{vardial2017report}. In the following, we will first describe our methodology and the data sets that we have used, before jumping to the results and some discussions in relation to our main findings.




\section{Methodology and Data}

\begin{figure*}[t]
    \centering
    \vspace{-0.3cm}
    \begin{tabular}{p{0.23\textwidth} p{0.02\textwidth} p{0.35\textwidth} p{0.02\textwidth} p{0.23\textwidth}}
      \vspace{0pt} \includegraphics[width=.23\textwidth]{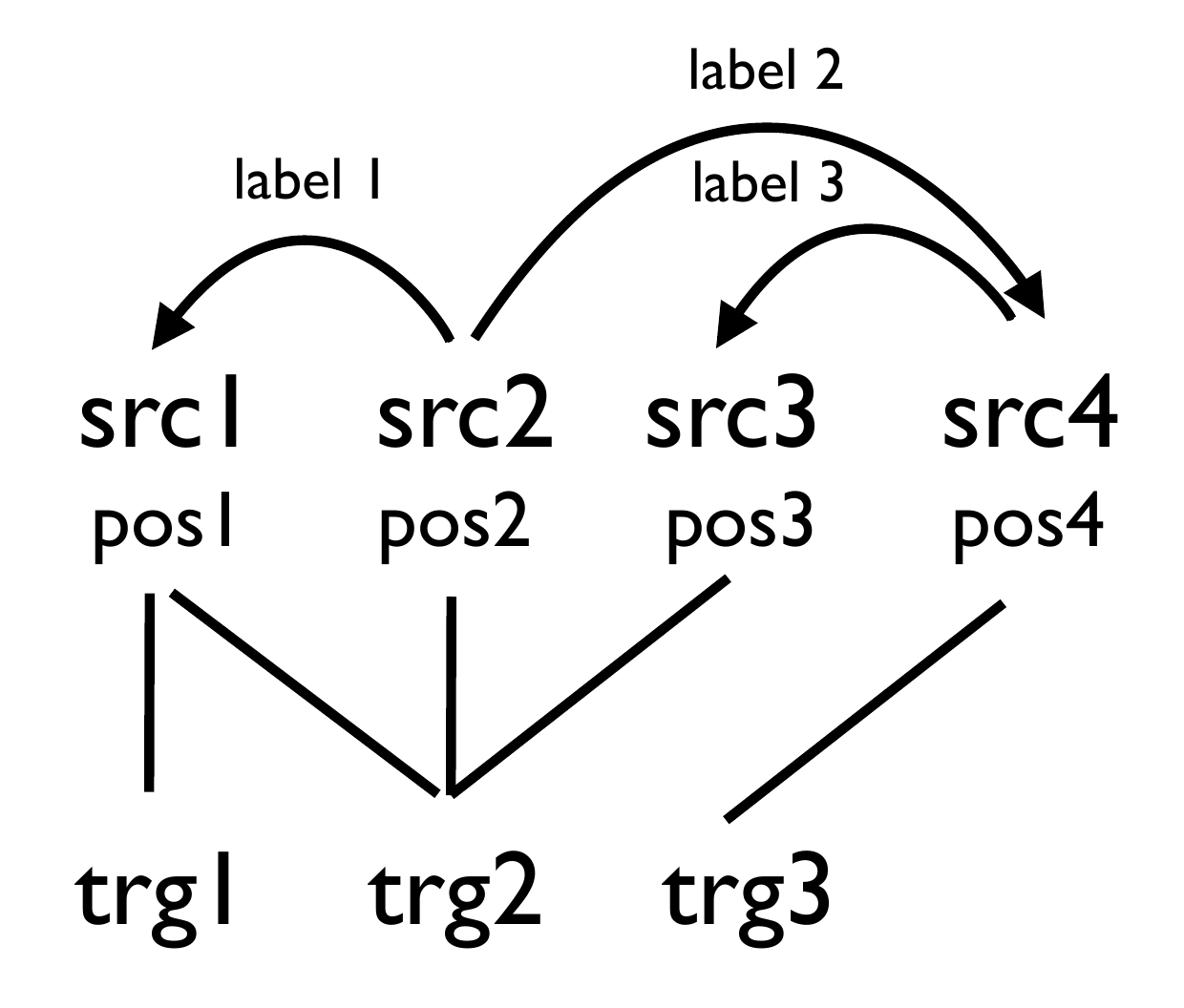}  & \vspace{2cm}{\Large\bf $\rightarrow$} &
      \vspace{0pt} \includegraphics[width=.35\textwidth]{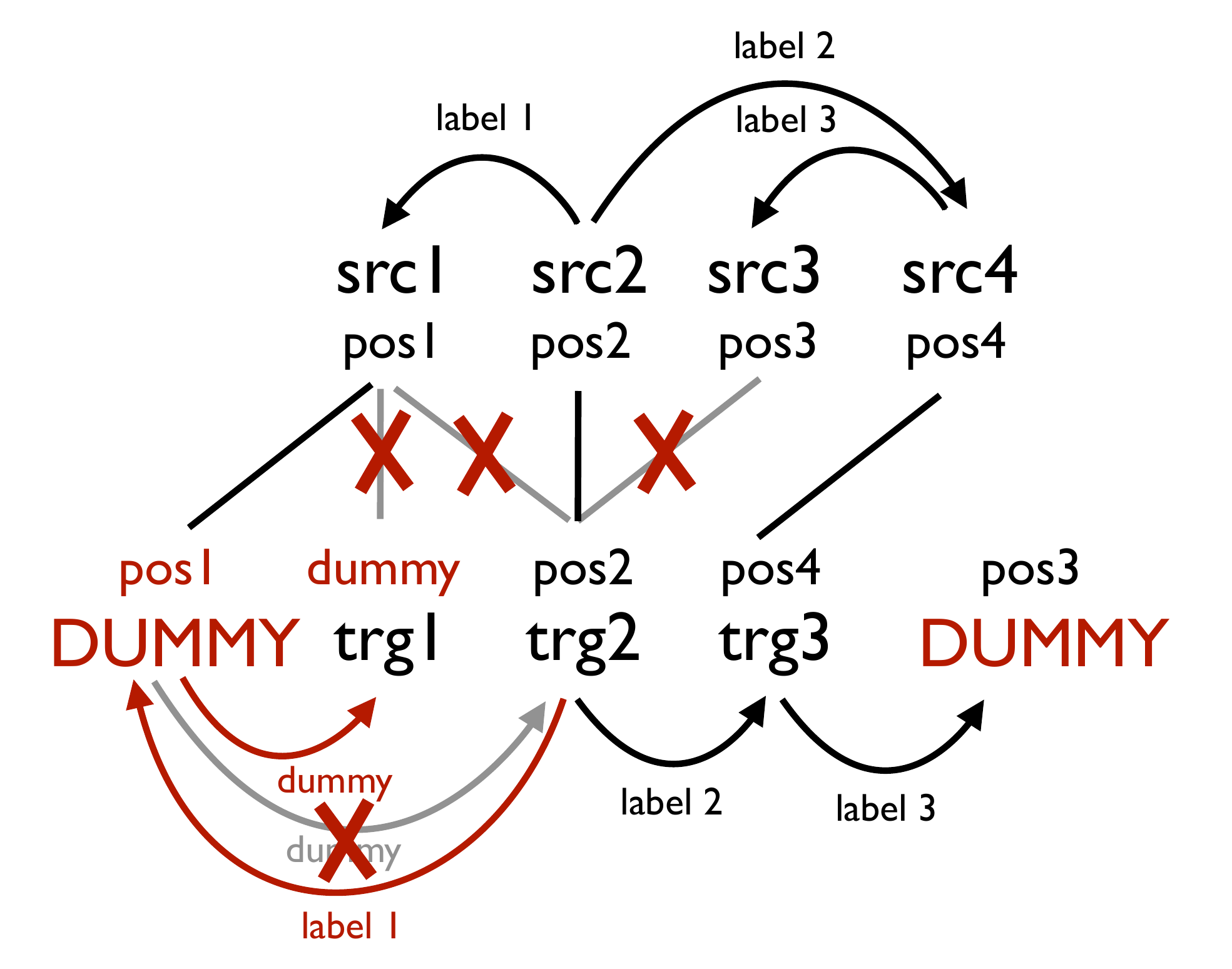}  & \vspace{2cm}{\Large\bf $\rightarrow$} &
      \vspace{0pt} \includegraphics[width=.23\textwidth]{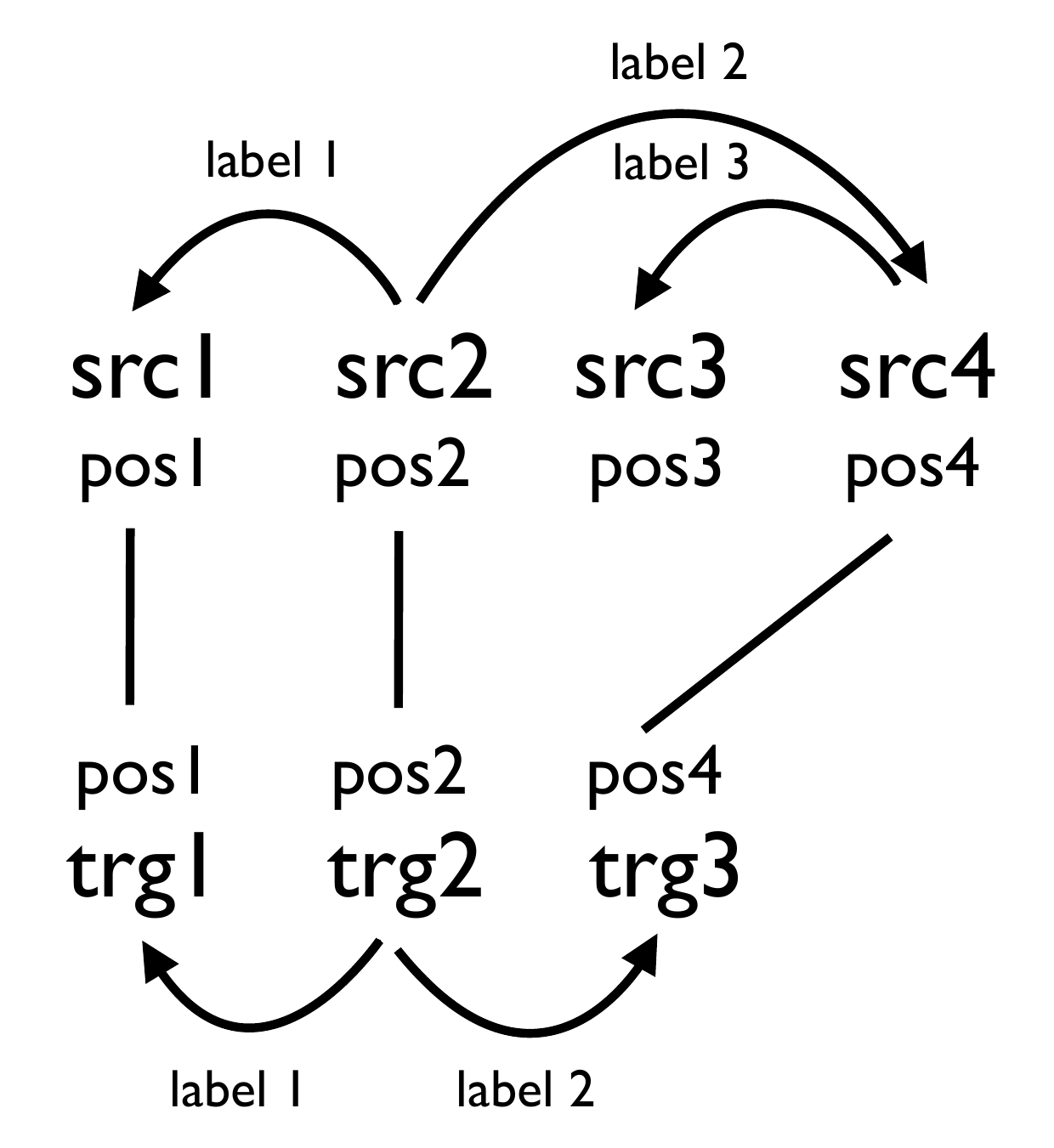} 
    \end{tabular}
    \vspace{-0.2cm}
    \caption{Annotation projection heuristics with {\em dummy} nodes: One-to-many alignments create dummy nodes that govern the linked target language tokens. Many-to-one alignments are resolved by removing links from lower nodes in the source language tree. Non-aligned source language tokens are covered by additional dummy nodes that take the same incoming and outgoing relations. The final picture to the right illustrates that dumme leaf nodes can safely be deleted and internal dummy nodes with single daughters can be removed by collapsing relations.}
    \label{img:nodummies}
  \end{figure*}

Our submission is based on previous work and basically applies models and techniques that have been proposed by \cite{Hwa05bootstrappingparsers,tiedemann:2014:COLING,TiedemannEA:CoNLL14}. We made very little changes to the basic algorithms but emphasized a systematic evaluation of different methods and parameters which we tested on the development data provided by VarDial 2017. All our results are, hence, scored on data sets with predicted PoS labels. In particular, we used three different cross-lingual model types:

\begin{description}
\item[Projection: ] Annotation projection across word-aligned parallel corpora using the data sets provided by the workshop organizers (the subtitle corpora). Source language parts are tagged and parsed automatically with supervised taggers and parsers.
\item[PBSMT: ] Treebank translation using a phrase-based model of statistical machine translation (SMT). Annotation are then projected from the original source language treebank to the translations to create a synthetic target language treebank. Alignment is taken directly from the translation model. The translation and language models are trained on the provided parallel corpora. No extra resources are used.
\item[SyntaxSMT:] Treebank translation using a tree-to-string hierarchical SMT model. Dependencies are transformed into constituency representations using the spans defined by the yield of each word in the sentence with respect to the dependency relations. Constituency labels are taken from the dependency relations and PoS labels are used as well for the labels of leaf nodes. After translation, we project the annotation of the source language treebank using the same procedures as for the other two approaches. Translation models are trained on the provided parallel corpora with automatically parsed source language sentences.
\end{description}

There are various improvements and heuristics for the projection of dependency trees. We applied two of them: (1) {\em collapseDummy}, which deletes leaf nodes that are labeled as ``dummy'' and also removes dummy nodes with just one daughter node by collapsing the parent and daughter relations. (2) {\em noDummy}, which discards all sentences that still include dummy nodes after applying collapseDummy.  Dummy nodes appear with the projection heuristics introduced by \cite{Hwa05bootstrappingparsers}, which we also use for handling non-one-ot-one word alignments. 
For example, unaligned source language tokens are projected on dummy target nodes to ensure the proper connection of the projected dependency tree. This can lead to dummy leaf nodes that can be ignored or dummy nodes with single daughters, which can be removed by collapsing the relations to head and dependent.  Figure~\ref{img:nodummies} illustrates the projection heuristics and the {\em collapseDummy} procedures. More details and examples are given in \cite{TiedemannAgic:2016:JAIR}. Overall, {\em noDummy} leads to a drop in performance and, therefore, we do not consider those results in this paper. The differences were small and most problems arose with smaller data sets where the reduction of training data has a negative impact on the performance.

For annotation projection, we used various sizes of parallel data to test the impact of data on parsing performance. The translation models are trained on the entire data provided by VarDial. Language models are simply trained on the target side of the parallel corpus.

We also tested cross-lingual models that exploit language similarities on the lexical level without translating or projecting annotation. The idea is similar to delexicalized models that are trained on generic features on the source language treebank, which are then applied to the target languages without further adaptation. With closely related languages, we can assume substantial lexical overlaps, which can be seen at the relative success of the second baseline in the shared task (also shown in Table~\ref{tab:basic-results}). In particular, we used substrings such as prefixes (simulating simple stemming) and suffixes (capturing inflectional similarities) to add lexical information to delexicalized models. However, those models did not perform very well and we omit the results in this paper.

For training the parsers, we used mate-tools \cite{bohnet2010top}, which gave us significantly better results than UDPipe \cite{udpipe:2016} without proper parameter optimization except for some delexicalized models. Table~\ref{tab:basic-results} compares the baseline models with the two different toolkits. We still apply UDPipe for PoS and morphological tagging using the provided tagger models for the target languages and similar ones trained on the UD treebanks for the source languages except for Czech, which did not work with standard settings due to the complexity of the tagset and limitations of the implementation of UDPipe. Instead, we apply Marmot \cite{mueller2015}  for Czech, which also provides efficient model training for PoS and morphology.

The only ``innovation'' compared to our previous work is the inclusion of target language tagging on top of annotation projection. Earlier, we only used projected annotation even for PoS information. In this paper, we also test the use of target language taggers (which are part of the provided setup) to (i) over-rule projected universal PoS tags and (ii) add morphological information to the data. Especially the latter makes a lot of sense especially for highly-inflecting languages like Slovak and Croatian. However, the risk of this procedure is that noisy projection of dependency label may not fit well together with the tags created by independent tools that are probably less noisy and make different kinds of mistakes. This may mislead the training algorithm to learn the wrong connections and we can see that effect in our experiments especially in connection with the tagging of universal PoS labels. This actually degrades the parsing performance in most cases. More details will be presented in the following section in connection with the results of our experiments.

\section{Results}
\label{sec:results}

We considered all language pairs from the VarDial campaign and here we present the relevant results from our experiments. First of all, we need to mention that we created new baselines using the mate-tools to have fair comparisons of the cross-lingual models with respect to baseline approaches. The new figures (on development data) are given in Table~\ref{tab:basic-results}. The same table also summarizes our basic results for all language pairs using the three approaches for data transfer as introduced in the previous section. All projections are made in collapseDummy mode as explained above.

\begin{table}[ht]
\centering
{\small
\begin{tabular}{|r|cccc|}
\hline
Target            & Croatian  & Slovak   & \multicolumn{2}{c|}{Norwegian} \\
Source           & Slovenian & Czech    & Danish & Swedish \\
\hline
\hline
{\em UDPipe}           &&&&\\
supervised     & 74.27       & 70.27    & \multicolumn{2}{c|}{78.10}\\
delex             & 53.93       & 53.66    & 54.54 & 56.71\\
cross             & 56.85       & 54.61    & 54.11 & 55.85\\
\hline
\hline
{\em mate-tools}    &&&&\\
supervised     & 79.68       & 71.89    & \multicolumn{2}{c|}{81.37}\\
delex             & 53.39       & 55.80    & 50.07 & 56.27\\
cross             & 60.29       & 62.21    & 56.85 & 59.63\\
\hline
\hline
{\em Projected}       & & & & \\
   100,000     &  58.82      & 60.29     & 57.19   & 63.03\\
   500,000     &  59.86      & 62.23     & 57.58   & 64.61\\
1,000,000     &  {\bf  62.92}      & 63.57     & 57.82   & 64.59\\
\hline
PBSMT           &  60.81      & {\bf 65.97}     & 57.87   & 65.96\\
\hline
SyntaxSMT    &  58.57      & 63.13     & 58.36   & {\bf 66.31}\\
\hline
\end{tabular}
}
\caption{Basic results of cross-lingual parsing models in terms of labeled attachment scores (LAS) on development data: Annotation projection on automatically parsed bitexts of varying sizes (projected: number of sentence pairs); treebank translation models (PBSMT and SyntaxSMT); compared to three baselines: delexicalized models (delex), source language models without adaptation (cross) and fully-supervised target language models (supervised).}
\label{tab:basic-results}
\end{table}

The first observation is that all cross-lingual models beat the delexicalized baseline by a large margin. This is, at least, self-assuring and motivates further developments in the direction of annotation projection and treebank translation. Another observation is that Croatian is surprisingly hard to improve in comparison to the cross-lingual model that applies a parser for Slovenian without any adaptation. 

Another surprise is the quality of the Norwegian models coming from Danish. Both languages are very close to each other especially in writing (considering that we use bokm{\aa}l in our data sets for Norwegian). Projection and translation should work well and should at least be on-par with using Swedish as the source language. However, the differences are quite significant between Danish and Swedish as the source language and this points to some substantial annotation differences between Danish and the other two languages that must be the reason behind this mystery. This conclusion is even more supported by the results of the cross-lingual baseline model without adaptation, which should perform better for Danish as the lexical overlap is large, greater than the overlap with Swedish. Yet another indication for the annotation differences is the result of the delexicalized parsers. There is also a big gap between Danish and Swedish as the source language. The result of these experiments demonstrate the remaining difficulties of cross-linguistically harmonized data sets, which is a useful outcome on its own.

We can also see, that treebank translation works rather well. For most language pairs, the performance is better than for annotation projection but the differences are rather small in many cases. An exception is Croatian for which annotation projection on parallel corpora works best, whereas translation is on par with Slovenian models applied to Croatian data.

In contrast to our previous findings, we can also see that the amount of data that is useful for annotation projection is bigger. Our prior work indicated that small corpora of around 40,000 sentence pairs are sufficient and that the learning curve levels out after that \cite{TiedemannAgic:2016:JAIR}. In this paper, we see increasing model performance until around one million sentence pairs before the scores converge (additional runs confirm this, even though they are not reported in the paper). A reason for this behaviour is that we now rely on movie subtitles instead of sentences from the European parliament proceedings. Subtitles are shorter in general and the domain may be even further away than parliament data, which explains the increased amount of data to obtain reasonable lexical coverage.

Our next study looks at the impact of tagging the target language with supervised models. Our previous work on annotation projection and treebank translation relied entirely on annotation transfer from source to target when training target language parsing models. This means that we discarded any language-specific features and modeled parsing exclusively around universal PoS tags and lexical information. For highly-inflecting languages, this is not very satisfactory and the performance drops significantly compared to models that have access to morphological features. Therefore, we now test models that use projected data with additional annotation from automatic taggers.
Table\ref{tab:add-tagging} summarizes the results of those experiments.

\begin{table}[ht]
\centering
{\small
\begin{tabular}{|l|ccc|}
\hline
 &           projected    &    \multicolumn{2}{c|}{target-tagged}\\
 &             PoS            &           morph   & PoS+morph\\
\hline
Projected & & & \\
sl-hr    & {\bf 62.92}    &      62.10 &       56.42      \\
cs-sk   & 63.57     &     --       &      70.68\\
da-no   &57.82       &   58.08    &     61.40       \\ 
sv-no   &64.59        &  64.78     &    62.35        \\
\hline
PBSMT & & & \\
sl-hr    &60.81 &        61.60 &          61.10   \\    
cs-sk   &67.81  &         --     &         {\bf 73.90}\\
da-no  &57.87    &    58.46     &       63.67        \\
sv-no  & 65.96     &  66.44      &       64.15        \\
\hline
SyntaxSMT & & & \\
sl-hr   & 58.57  &       60.15   &        56.85   \\    
cs-sk  & 63.13   &     64.05      &       65.02\\
da-no  &58.36     &   58.59       &     64.74        \\
sv-no  &66.31      &  66.64        &     65.43        \\
da+sv-no & --     & \multicolumn{2}{c|}{{\bf 67.80}} \\
\hline
\end{tabular}
}
\caption{Added PoS and morphological tagging to projected data sets: LAS scores on development data. Only morphological tagging added (morph) or tagging both, PoS and morphology (PoS+morph).}
\label{tab:add-tagging}
\end{table}

There are two models that we evaluate: (i) A model that adds morphological features to the projected annotation, and (ii) a model that even overwrites the universal PoS tags created through projection. The first variant adds information that may contradict the PoS labels transferred from the source. For example, it may assign nominal inflection categories to a word labeled as verb through projection. The latter model should be more consistent between PoS and morphology but has the problem that those categories may not fit the dependency relations attached to the corresponding words when projecting from the source. This can also greatly confuse the learning procedures.

As it turns out, overwriting the projected PoS labels is more severe in most cases except Slovak and Norwegian (only when projected from Danish). There, it seems to be beneficial to run complete tagging after projection. In almost all other cases the performance drops, often quite dramatical. On the other hand, adding morphology always helps, except for Croatian annotation projection (which is a bit surprising again).

There is no clear winner between phrase-based and syntax-based SMT. For Slovak and Croatian, phrase-based systems seem to work best whereas Norwegian performs better with syntax-based models. A combination of Danish and Swedish data gives another significant boost (retagging projected Danish including PoS and adding morphology to projected Swedish).

We then used the best results on development data for each of the three target languages to run the cross-lingual models on the test set. No further adjustments were done after tuning the models on development data. The final results of the official test are shown in Table~\ref{tab:final-result}.

\begin{table}[ht]
\centering
{
\begin{tabular}{|l|ccc|}
\hline
LAS               &     hr  &   no    &     sk\\
\hline
\hline
supervised   &73.37 &  81.77  & 71.41\\
delex            &50.05 & 58.13   & 53.87\\
cross            &56.91 & 60.22   & 61.17\\
\hline
\hline
CUNI             & 60.70 & 70.21 & 78.12\\
our model     & 57.98 & 68.60  &73.14\\
\hline
\end{tabular}

\bigskip

\begin{tabular}{|l|ccc|}
\hline
UAS               &     hr  &   no    &     sk\\
\hline
\hline
supervised   &80.16 & 85.59  & 78.73\\
delex            & 63.29 & 67.86   & 64.55\\
cross            & 68.52 & 69.31   & 70.60\\
\hline
\hline
CUNI             & 69.73 & 77.13  & 84.92\\
our model    & 69.57 & 76.77  &82.87\\
\hline
\end{tabular}
}
\caption{Final results on the test set ({\em our model}) compared to baselines and fully supervised models. {\em CUNI} refers to a competing system -- the winning team of VarDial. For the Norwegian baselines we report the results for Swedish as the source language, which is much better than using Danish.}
\label{tab:final-result}
\end{table}

The results on test data mainly confirm the findings from the development phase. Slovak performs clearly best in the cross-lingual scenario. This is the only language pair for which the cross-lingual model even outperforms the fully supervised ``upper bound''. This is quite fascinating and rather unexpected. Certainly, the Czech treebank is by far the largest one in the collection and much bigger than the corresponding Slovak treebank. The languages are also very close to each other and their morphological complexity requires sufficient resources. This may explain why the large Czech training data can compensate for the shortcomings of the small Slovak training data. Other factors for the positive result may also include the similarity in domains covered by both treebanks and the closeness of annotation principles.
The performance for the other target languages is less impressive.
Norwegian performs similar to the scores that we have seen in related work on annotation projection and cross-lingual parsing. Croatian is rather disappointing even though it also beats the cross-lingual baselines.

The main scores in our evaluations is LAS but it is also interesting to look at unlabelled attachment scores (UAS). Table~\ref{tab:final-result} lists those scores as well and we can see that labelling seems to be a major problems for our models. The difference to LAS scores is dramatic, much more than the absolute difference we see between UAS and LAS in the fully supervised models. Compared to the winning submission at VarDial ({\em CUNI}, see \cite{rosa2017dsl}), we can also see that the main difference is in LAS whereas UAS are rather similar. This seems to be a shortcoming of our approach that we should investigate more carefully.

\section{Conclusions}


Our experiments demonstrate the use of annotation projection and treebank translation techniques. The  models perform well, especially for Slovak, which even outperforms the fully supervised ``upper bound'' model. In this paper, we have discussed the use of target language tagging on top of annotation projection with the conclusion that adding morphological information is almost always useful. We observe a large gap between LAS and UAS, which would require some deeper investigations. A possible reason is the use of language-specific dependency labels that are not available from the projection. However, we actually doubt that explanation looking at the success of the winning team. In their results, LAS did not suffer that much. Some surprising results could be seen as well, for example, the fact that Danish does not work as well as a source for Norwegian as Swedish does. This cannot be explained in terms of linguistic grounds but need to refer to unexpected annotation differences or possibly a larger domain mismatch. Croatian as a target language was also surprisingly difficult and the performance is the worst in the final among all test cases. This improvement over the non-adapted Slovenian parser is only very modest whereas large gains can be observed for the other language pairs.

\bibliography{vardial2017}
\bibliographystyle{eacl2017}

\end{document}